\documentclass{article}

\usepackage{arxiv}

\usepackage[utf8]{inputenc} 
\usepackage[T1]{fontenc}    
\usepackage{url}            
\usepackage{booktabs}       
\usepackage{nicefrac}       
\usepackage{amsmath}
\usepackage{amssymb}
\usepackage{graphicx}
\usepackage{natbib}
\usepackage{doi}
\usepackage{latexsym}
\usepackage{xcolor}
\usepackage{subfig}
\usepackage{listings}
\usepackage{etoc}
\usepackage{caption}
\usepackage{floatrow}
\usepackage{etoolbox} 

\patchcmd{\thanks}{#1}{\protect\doublespacing\doublespacing\normalsize#1}{}{}

\title{Equitable modelling of brain imaging by counterfactual augmentation with morphologically constrained 3D deep generative models}

\author{Guilherme Pombo \thanks{Corresponding author: guilherme.pombo.18@ucl.ac.uk} \thanks{UCL Queen Square Institute of Neurology, University College London, London, UK} \and Robert Gray \footnotemark[2] \and M. Jorge Cardoso \thanks{School of Biomedical Engineering \& Imaging Sciences, King's College London, London, UK} \and Sebastien Ourselin \footnotemark[3] \and Geraint Rees \footnotemark[2] \thanks{Faculty of Life Sciences, University College London, London, UK}  \and John Ashburner \footnotemark[2] \and Parashkev Nachev \footnotemark[2] }

\renewcommand{\headeright}{CounterSynth}
\renewcommand{\shorttitle}{G.Pombo et al.}

\hypersetup{
pdftitle={Equitable modelling of brain imaging by counterfactual augmentation with morphologically constrained 3D deep generative models},
pdfsubject={q-bio.NC, q-bio.QM},
pdfauthor={Guilherme Pombo, Robert Gray, M. Jorge Cardoso, Sebastien Ourselin, Geraint Rees, John Ashburner, Parashkev Nachev},
pdfkeywords={Counterfactual synthesis, Deep generative models, Discriminative models, Data augmentation,
Fairness, Equity, Brain imaging},
}

\begin{document}
\maketitle

\begin{abstract}
We describe Countersynth, a conditional generative model of diffeomorphic deformations that induce label-driven, biologically plausible changes in volumetric brain images. The model is intended to synthesise counterfactual training data augmentations for downstream discriminative modelling tasks where fidelity is limited by data imbalance, distributional instability, confounding, or underspecification, and exhibits inequitable performance across distinct subpopulations.

Focusing on demographic attributes, we evaluate the quality of synthesized counterfactuals with voxel-based morphometry, classification and regression of the conditioning attributes, and the Fr\'{e}chet inception distance. Examining  downstream discriminative performance in the context of engineered demographic imbalance and confounding, we use UK Biobank magnetic resonance imaging data to benchmark CounterSynth augmentation against current solutions to these problems. We achieve state-of-the-art improvements, both in overall fidelity and equity. The source code for CounterSynth is available online.
\end{abstract}


\section{Introduction}
The manifestations of neurological disease in the imaged brain are complex, reflecting the intersection of pathological, biological and instrumental forms of variation. A signal of interest here must typically be disentangled from a rich, widely distributed network of interacting factors: some irrelevant, others modulating. This problem is traditionally approached by assuming an \textit{a priori}-defined, simple underlying compositionality --- into discrete anatomical regions or continuous stereotactic spaces --- that enables compact models to be deployed in a regional or voxel-wise manner \citep{vbm_intro, anatomical_svm}. So strong a simplifying assumption places a hard limit on the complexity of the signals that can be modelled, but is inevitable where the scale of available data is small and the controllable flexibility of the models fitted to it low.

The revolutionary impact of deep learning on image modelling \citep{alexnet, resnet, human_level} may enable us to relax this assumption \citep{brain_disease_diagnosis, brain_disease_diagnosis_2, pombo, tumor_segmentation}. Given sufficiently informative data, a deep neural network can implicitly find a decomposition of the image that best supports the task it is deployed to solve --- prediction, prescription, or inference --- trained end-to-end, guided by only weak inductive bias. Though attainable model expressivity is thereby enhanced, it falls on the data to control it. Crucially, any model here must rely on the data to distinguish between the target, foreground \textbf{signal}, and incidental, background \textbf{context} in which it is embedded.

Simple forms of context independence, such as translational \citep{lecun_rotations} and rotational \citep{spherical_cnns} invariance, or approximate viewpoint invariance \citep{capsule_networks} can be incorporated in the model's design. Equally, approximate invariance to geometric and intensity transformations can be promoted with on-the-fly data augmentation: this is how most deep image classifiers learn invariance to, for example, (small) affine and elastic transformations \citep{info11020125, survey_on_data_augmentation} and --- in the context of medical imaging --- bias field and motion artefacts \citep{MONAI}. Models can even learn which augmentations to learn \citep{learning_invariances}.

Where the context of a predictive signal is \textit{itself} complex --- for example, the age-related morphology of a brain in which small vessel disease is the target --- no simple remedy is available. Retained sensitivity to context here not only impairs fidelity, it introduces vulnerability to distributional shifts, and may inject bias through irrelevant natural (confounder) \citep{gebru_gender, fairness, gender_imbalanced_xray} or sampling (collider) correlations \citep{griffith2020collider, watson2019collider}. The class imbalance and inadequate data representation common in the clinical domain \citep{clinical_disparity, google_under}  can only amplify the risks.

These concerns are far from merely hypothetical. Small vessel disease and age-related involutional change are closely correlated yet causally distinct \citep{small_vessel_age}. A naive model could easily learn to rely on age \citep{intriguing, out_of_dist} to support an inference on small vessel disease, resulting in impaired performance in the decision space of highest clinical significance: where the two are unusually decorrelated. Similarly, a model tasked with distinguishing ischaemic from inflammatory causes of white matter hyperintensity (WMH) will be drawn into favouring the former in elderly men and the latter in young women \citep{sex_diff_neuroinflammation, bonkhoff2021female}, reflecting the marked interactions of age and sex in the underlying patterns of disease prevalence. Many other examples are easy to adduce \citep{predictive_modelling, hiv_chained, chinese_brain, gender_imbalanced_xray}: the entanglement of pathological signals of interest with background, contextual factors is here not the exception, but the norm. 

These concerns are also ethical. Amongst the many contextual factors in play are those --- such as age, sex and ethnicity --- that define demographic subpopulations to which model performance and downstream clinical care ought to be invariant. It is not merely the mean fidelity of the model but its equity across heterogeneous populations that matters, a broader notion of fidelity inherent in the fundamental nature of medicine. Contextual invariance here must not only be implicitly promoted but explicitly demonstrated. 

The space of possible solutions to this cardinal problem is dominated by two distinct approaches. One is to redistribute the model's attention in training, through targeted data weighting or resampling \citep{undoing_dataset_bias, importance_weighting}, context-dependent modulation of the objective \citep{dro_neural_nets}, or adversarial mechanisms \citep{confounder_free}. The redistributive nature of these approaches tends to incur a penalty in model fidelity, even if generalisation or equity may be improved, and is limited to the maximum disentanglement that the available training data supports. 

An alternative is to augment the training data with samples from a generative model expressive enough to capture the interactions between the target signal and its context, in direct evolution of the use of generative models to expand minority classes synthetically \citep{gan_min_oversampling, capsule_oversampling}. Though this approach is theoretically superior, its success is premised precisely on the disentanglement we are using it to promote. Conversely, the successful transfer of a deep generative model that has learnt a contextual feature such as age from another data set is impeded by its ignorance of the target pathology, which it will tend to erase.

Here we propose a way out of this impasse through the use of conditional generative models of diffeomorphic (spatial) deformations \citep{dartel, template_construction, unsupervised_diffeo, dalca_voxelmorph}. Constraining the synthesis of an image to the diffeomorphic deformation of another provides the flexibility to capture common background morphological patterns of contextual modulation \citep{ageing_patterns, sex_morphology},  while leaving the target pathological signals relatively intact. Deformation fields can also be synthesised and re-sampled quickly, enabling on-the-fly augmentation even at very high image resolutions.

Moreover, a model of deformation alone is less complex and therefore easier to fit in the limited data regimes common in brain imaging.

Our general solution to the problem of promoting contextual invariance in models of brain imaging, \emph{improving both model fidelity and equity}, is summarised as follows:

\begin{itemize}
    \item We describe a fully-3D conditional generative model, CounterSynth, that synthesises counterfactual volumetric brain imaging for targeted, biologically informed augmentation of downstream discriminative models. Synthesis involves sampling a diffeomorphic deformation field conditioned on an original image and a contextual variable of interest, such as age or sex.
    \item  Synthesis modifies only select morphological features of the source volume, presenting the target pathological signal against alternative, prescribed, counterfactually defined backgrounds. The deformation fields are easily regularised to promote minimal, biologically plausible deformations, even when conditioning on abnormal images. Modelling shape, but not signal, further enhances the robustness of the generative model to natural variations in signal intensity.
    \item CounterSynth is fast and memory efficient: it can generate training augmentations at sub-millimeter resolutions on-the-fly, even on consumer-grade hardware. The deformations can be resampled with negligible cost for fast synthesis at multiple resolutions.
    \item We use synthesised counterfactuals to mitigate the impact of demographic imbalance, spurious correlations, and collider bias on brain imaging classification and regression tasks.
    \item  We quantify the value --- to both overall fidelity and equity --- of augmenting data with synthesised counterfactuals, in comparison with confounder-free networks (see \S\ref{cf_net_section}) and in comparison and combination with group distributionally robust optimisation (see \S\ref{group_dro_section}), demonstrating superiority to current practice on both counts. In the course of this evaluation we introduce novel indices of equitable model performance and its cost.
    \item Our code and trained models are available online.
\end{itemize}

\subsection{Related work}\label{related_work}
An image classifier is a function that assigns each observation in image-space, $\mathcal{X}$, a label in label-space, $\mathcal{Y}$. Suppose we are given a family, $\Theta$, of image classifiers, a loss (risk) function $\ell: \Theta \times(\mathcal{X} \times \mathcal{Y}) \rightarrow \mathbb{R}$, and $N$ image-label pairs $(x_1,y_1),\dots,(x_N,y_N)\in\mathcal{X}\times\mathcal{Y}$. The usual approach to model selection, which is based on empirical risk minimization (ERM), is to find a classifier $\theta \in \Theta$ that minimizes the empirical loss (risk), $\frac{1}{N}\sum_{k=1}^N\ell(\theta, (x_k, y_k))$. If the observations are not sufficiently homogeneous, however, then prioritizing \emph{average} performance can lead to important subgroups being underserved \citep{gebru_gender,  dro_neural_nets}.

\subsubsection{Group distributionally robust optimisation}\label{group_dro_section}
In distributionally robust optimization (DRO) \citep{ben2013robust, rahimian2019distributionally}, one aims to minimize the worst-case expected loss over an `uncertainty set' of distributions. In the group DRO setting \citep{does_dro_work, dro_language, dro_neural_nets}, this minimisation is simply over the (instantaneous) worst-performing group of examples. In the context of neural network optimisation, given training data already divided into groups, \citep{dro_neural_nets} minimise this empirical worst-group risk while demonstrating the importance of simultaneously enhancing generalisability through greater regularisation. They achieve markedly improved worst-group test set accuracy over ERM-based approaches, with minimal reductions in average test set performance. We benchmark CounterSynth against, and in conjunction with, group DRO. In particular we show that we can improve worst-group performance without harming average performance.

\subsubsection{Unpaired image-to-image translation and GANs}\label{unpaired_image_to_image_translation}
Unconditional generation of counterfactual volumetric brain imaging is as challenging as our ultimate objective of equitable predictive modelling. Hence our focus is (unpaired) image-to-image translation, whereby a given image is transferred to a new `domain' by a conditional generative model.

The state of the art here, StarGAN (v1 \cite{stargan}; v2 \cite{stargan2}), is a type of generative adversarial network (GAN), \citep{gan_goodfellow, cycle_gan, stargan}. These two-part neural networks comprise a `generator', G - a neural network that maps a random vector to image space - and a `discriminator', D - a neural binary classifier that distinguishes between training data and the generator's output. Given the data distribution $p(X)$ and a latent distribution $p(Z)$, the models train simultaneously by playing the two-player minimax game
\begin{equation}
    \label{gan_objective}
\mathop{min}_{G}\mathop{max}_{D} \mathbb{E}_{x\sim p(X)} \log D(x) + \mathbb{E}_{z\sim p(Z)}\log(1-D(G(z))).
\end{equation}
Under various technical conditions \citep{gan_goodfellow, stability_of_gans, r1_regularisation} the distribution $G(z)$, $z\sim p(Z)$, converges to $p(X)$.

StarGAN is a \emph{conditional} GAN in the sense that, instead of noise, the generator takes as input an image and one or more domain labels. Multiple domain labels enables simultaneous transfer between multiple domains (e.g. when modelling portraits changing hair colour \emph{and} facial expression), while also allowing a subset of image properties to be fixed while others are transformed (e.g. adding spectacles while keeping age constant).

At StarGAN's core is a type of `cycle consistency loss' \citep{cycle_gan}, an L1 penalty on the reconstruction error accumulated by transferring an image to a domain and back again. In terms of the joint distributions of images and labels, $p(X,Y)$, and the marginal label distribution $p(Y)$, their cycle loss is
\begin{equation}\label{cycle_loss}
\mathbb{E}_{(x,y)\sim p(X,Y), y_\text{new}\sim p(Y)} ||x - G( G(x, y_\text{new}), y)||_1.
\end{equation}
This loss encourages approximately invertible domain transfers that, when regularised, should be no more complex than necessary. Domain transfers are thereby encouraged to preserve the visual content of the original image.

\subsubsection{Confounder-free neural network}\label{cf_net_section}
The confounder-free neural network (CF-Net) learning scheme \citep{confounder_free} is designed to discourage medical image prediction models from acquiring biases in the presence of confounders. A minimax-type adversarial objective \eqref{gan_objective} is used to promote approximate invariance of the predictor's featurisation of the image data to the presence of a given confounder in the input. The method has been validated on several challenging real-world diagnosis prediction tasks, including prediction of human immunodeficiency virus status from brain imaging data. We compare CF-Net to models trained on CounterSynth synthetic counterfactuals in \S\ref{demographically_fair_predictions}.

\subsubsection{Volumetric synthesis with GANs}\label{volumetric_modelling_with_GANs}
StarGAN v2 is trained and tested on images of $256\times256$ resolution; our (volumetric) imaging dimensionality is greater by a factor of 32 (see \S\ref{the_data}). Though 3D-StyleGAN \citep{3d_style_gan} shows that GANs are capable of generating realistic volumetric data unconditionally at $4$ times the dimensionality of the original, $64 \times 64 \times 64$, its sampling failures (see p8 of \cite{3d_style_gan}) suggest our more structured approach to prediction is warranted.

Pseudo-3D models have lower computational complexity. For example, 4D-DANI-Net \citep{dani_net} learns from matched pairs of volumes to perform domain transfer on Alzheimer's disease imaging. An alternative model, ADESyn \citep{ade_syn}, does not require matched imaging. Both of these methods incorporate mechanisms for recovering some of the spatial correlations lost in a pseudo-3D framework.

The computational burden can also be reduced by training on small patches. A conditional super-resolution model that operates on $64\times40\times64$ subvolumes is described in \citep{wang2020enhanced}. Similarly, the conditional model described in \citep{mri_superres_gan_2} operates on $96\times96\times48$ regions of interest. An attractive property of CounterSynth is that its deformation fields can simply be cropped and resampled to operate on image patches of any shape.

\subsubsection{Equitable model performance}\label{machine_learning_fairness}
Medicine is concerned with minimizing the difference, at the individual level, between ideal and achieved clinical outcomes. Since the optimal management of an individual patient is typically unknown, it must be inferred from the population. In the setting of population heterogeneity, the fidelity of such inference will tend to be systematically biased in proportion to the representation of any given subpopulation \citep{gender_imbalanced_xray, gebru_gender, google_under}. The problem of equity then arises as consistent variation in model performance across different subpopulations.

Equity can be promoted by data manipulation prior to modelling \citep{survey_on_class_imbalance}, or by directly incorporating appropriate metrics into training objectives \citep{dro_language, dro_neural_nets, fairness_ml}. In rebalancing the model's attention across the population, any benefit to a given subpopulation may incur an undesirable cost elsewhere \citep{dro_neural_nets}.

In this paper, we therefore consider variations in model performance at both the local (subpopulation) level and global level, quantifying the improvements at a local level with regards to the changes to global performance (see \S\ref{cost_index}).

\section{Methodology}\label{methodology}
We use StarGAN-based unpaired image-to-image style transfer to synthesise realistic counterfactual brain imaging in terms of diffeomorphic deformations. These are infinitely differentiable, invertible coordinate transformations with infinitely differentiable inverses \citep{dartel, ashburner_longitudinal, template_construction}.

Our restriction to deformations has the benefits itemised in the introduction; the restriction to (regularised) diffeomorphisms in particular enables us to dispense with the cycle loss \eqref{cycle_loss} because regularised diffeomorphic displacements naturally produce simpler, invertible domain transfers. This considerably reduces model run time\footnote{Each evaluation of the cycle loss \eqref{cycle_loss} requires two forward passes through the generator, one of which is required to evaluate \eqref{gan_objective}, the other we now avoid.} as well as the complexity of the training objective.
We also forego the `style vectors' that were introduced in StarGAN v2 \cite{stargan2}) to further simplify the training objective; hence our model is closest to StarGAN v1.

\subsection{Learning diffeomorphic deformations}\label{learning_deformations}
Methods for predicting diffeomorphic deformations with neural networks by using `spatial transformer layers' \citep{spatial_transformers} are described independently by \citet{unsupervised_diffeo} and \citet{dalca_voxelmorph}. In both cases a convolutional neural network (CNN) predicts a coordinate transformation $\boldsymbol{\phi}:\mathbb{R}^3\mapsto\mathbb{R}^3$ that registers a given source volume onto a given target volume. The deformation $\boldsymbol{\phi}$ is represented in terms of a stationary velocity field, $\boldsymbol{v}$, a real parameter $t\in[0,1]$ and the identity transformation $\text{Id}$, defined such that
\begin{equation}
    \label{eqn:ode}
    \frac{\partial \boldsymbol{\phi}^{(t)}}{\partial t}=\boldsymbol{v}(\boldsymbol{\phi}^{(t)}),
    \qquad \boldsymbol{\phi}^{(0)}=\text{Id}.
\end{equation}
Integrating $t$ over $[0,1]$ or, equivalently, exponentiating $\boldsymbol{v}$ recovers the deformation: $\boldsymbol{\phi}=\boldsymbol{\phi}^{(1)}=\exp(\boldsymbol{v})$. Their CNNs predict $\boldsymbol{v}$ from the source and target images, and then integrate $\boldsymbol{v}$ by scaling and squaring \citep{square_scale, dartel, scaling_squaring} before finally applying $\boldsymbol{\phi}$ to the source image.

Crucially, all of the spatial transformations in this network are implemented in terms of (sub)differentiatiable `spatial transformer layers' \citep{spatial_transformers}, so that every operation is a (sub)differentiable function of its input. Therefore the entire model can be optimised end-to-end, simply using (sub)gradient descent.

In the following section we describe how we use this technique to predict deformations for counterfactual synthesis.

\subsection{Counterfactual synthesis with deformations}\label{diffeo_gan}
Our training objective is based on that of StarGAN \citep{stargan}. Suppose we have a set of domain labels $\{0,\cdots,N\}$ and let $\mathcal{U}$ denote the discrete uniform distribution over this set.
Using the notation from \citet{stargan}, we use a discriminator $D_{s r c}$ to classify images as training data or not training data. The main component of our objective is the `non-saturating' \citep{gan_goodfellow} alternative to \eqref{gan_objective}, which is used to encourage \emph{all} deformations to be realistic:
$$
\mathcal{L}_{a d v} = \mathbb{E}_{x\sim p(X)}\left[\log D_{s r c}(x) - \mathbb{E}_{\mathbf{c} \sim \mathcal{U}}\log D_{s r c}( \boldsymbol{x} \circ \boldsymbol{\phi}(\mathbf{x},\mathbf{c}) )\right].
$$
We use a second discriminator $D_{c l s}$ to predict the domain of an image, and we let $D_{c l s}\left(\mathbf{c} \mid \mathbf{x}\right)$ represent the probability distribution over domain labels predicted by $D_{c l s}$. In terms of the joint distribution of images and (true) domain labels, $p(X, Y)$, we minimise the following with respect to $D_{c l s}$,
$$
\mathcal{L}_{c l s}^{real} = -\mathbb{E}_{(\mathbf{x}, \mathbf{c}) \sim P(X, Y)} \log D_{c l s}\left(\mathbf{c} \mid \mathbf{x}\right).
$$
Given an image $x$, let $\bar{ \mathcal{U} }_x $ denote the uniform distribution over its counterfactual domain labels. To learn to generate counterfactuals, we minimise the following with respect to $\boldsymbol{ \phi }$,

\begin{equation}\label{learn_counterfactuals}
\mathcal{L}_{c l s}^{fake} = -\mathbb{E}_{ \mathbf{x} \sim P(X), \mathbf{c} \sim \bar{ \mathcal{U}_x} } \log D_{c l s}( \mathbf{c} \mid \mathbf{x} \circ \boldsymbol{\phi} ( \mathbf{x}, \mathbf{c} )).
\end{equation}

We smooth the velocity field $\boldsymbol{v}$ in $\boldsymbol{\phi}( \boldsymbol{x}, \boldsymbol{c} ) = \exp( \boldsymbol{v} ) $ as in \citep{dalca_voxelmorph}, by using a diffusion regularizer on its spatial gradients: for each voxel $(i, j, k)$ ,

\begin{equation}
    \mathcal{L}_{\text{smooth}} = \mathbb{E}_{(\mathbf{x}, \mathbf{c}) \sim P(X, Y)} \sum_{i, j, k} \left\| \nabla \boldsymbol{v} (\mathbf{x}, \mathbf{c}) (i, j, k) \right\|^{2}.
\end{equation}

Finally, for more stable training we use $R_{1}$ regularisation (\citet{stargan2}; see also \S4.1 of \cite{r1_regularisation}),
\begin{equation}
R_{1} = E_{\mathbf{x} \sim P(X) }\left\|\nabla D(\mathbf{x})\right\|^{2}.
\end{equation}
In summary, our objective is to minimise $\mathcal{L}_{disc}$ with respect to $D_{s r c}$ and $D_{c l s}$, while minimising $\mathcal{L}_{gen}$ with respect to $\boldsymbol{\boldsymbol{\phi}}$, where

\begin{align}
\mathcal{L}_{disc} & = -\mathcal{L}_{a d v} + \mathcal{L}_{c l s}^{real} + R_1, \nonumber\\
\mathcal{L}_{gen} & = \mathcal{L}_{a d v} + \mathcal{L}_{c l s}^{fake} + \mathcal{L}_{\text {smooth }}.\nonumber
\end{align}

\begin{figure*}
    \centering
    \includegraphics[width=16cm]{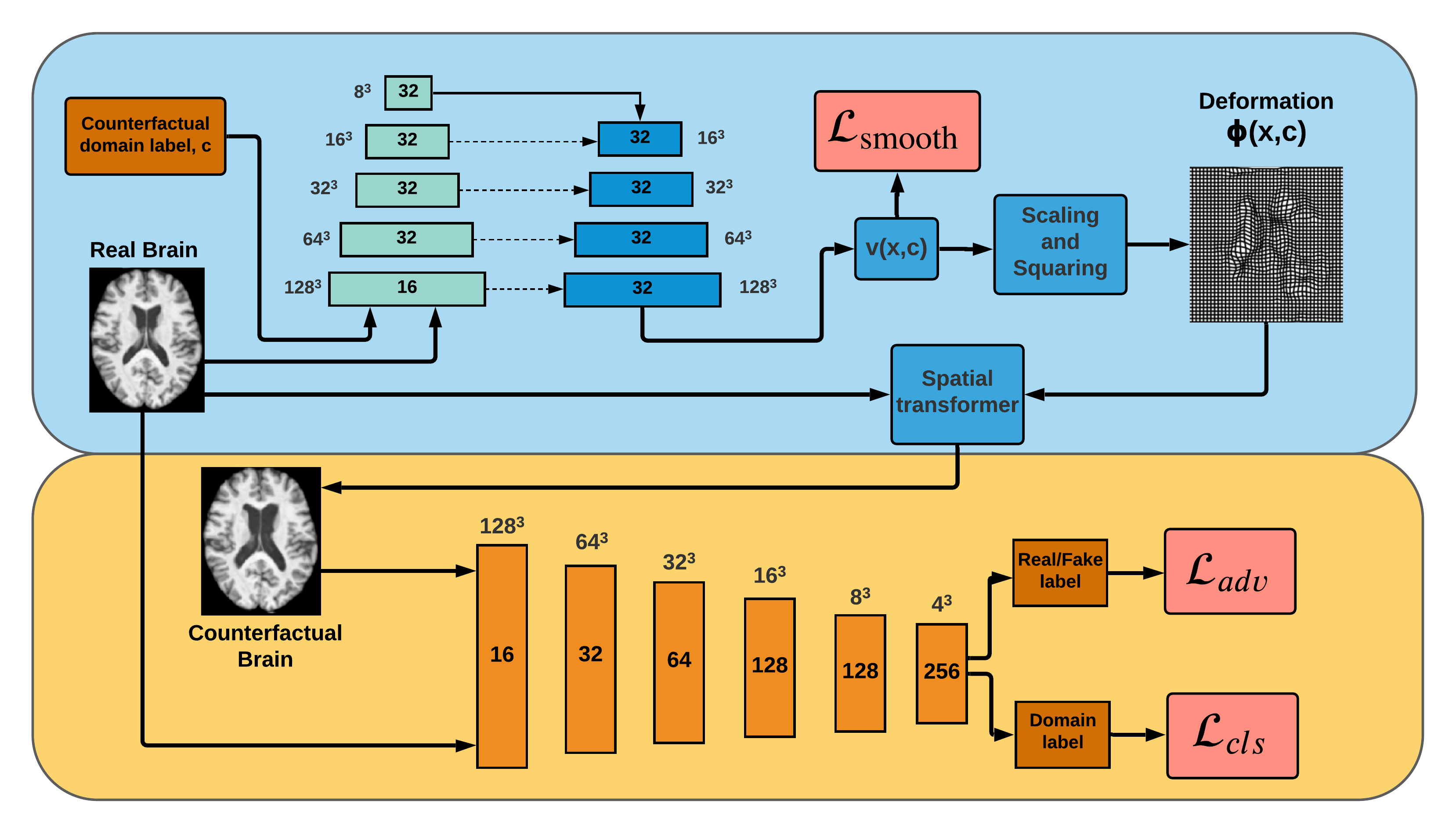}
     \caption[Gan architecture]{\textbf{Top}: The U-Net, plus scaling and squaring layers, for predicting and applying the deformation $\boldsymbol{\phi}$, via the velocity $\boldsymbol{v}$. The input is the real image together with the counterfactual label added as a second image channel. Each block in both pyramids of U-Net layers is a convolutional layer that produces a feature map with $16$, and thereafter $32$, channels. Next to each block is its spatial resolution. These resolutions are decreased with max pooling and increased with nearest neighbour resampling. Dotted arrows represent skip connections. The scaling and squaring block is composed of spatial transformer layers. \textbf{Bottom}: The fully-convolutional discriminator for classifying real and synthesised images. Each upright block is a convolutional layer, producing feature maps with $16,\dots,256$ channels. Above each block is the spatial resolution. Max pooling is used to reduce this resolution. Two probability distributions are predicted: real vs fake, and a distribution over domain labels.}
    \label{fig:total_gan}
\end{figure*}

\subsection{Quantifying equity of performance} \label{cost_index}
To quantify the impact of augmentation with counterfactual synthetic data on the relative equity of two classifiers (see \S\ref{machine_learning_fairness}), we would like a principled method of indexing variations in performance across the population. Econometrics provides an array of equality indices, such as the Gini index \citep{gini_coefficient}, the Theil index \citep{theil_index} and the concentration index \citep{concentration_index}, largely based on statistical measures of dispersion. 

In the context of model fidelity, equity can be trivially achieved by lowering global performance to that of the worst-performing subpopulation, but we typically wish to improve local performance without harming global performance. To capture the global impact of any \emph{local} intervention, we need to measure \emph{both} local and global effects. 

First, we divide the population into the subpopulations $A_1, \dots, A_N$, and denote by $a_k$ the mean performance of a given classifier on subpopulation $A_k$. Here we model global performance as the mean of these means: $\frac{1} {N}\sum_{k} a_k$. Of course, if each $A_k$ is equal in size this simplifies to the population mean. This approach weights every subpopulation equally, ensuring that its contribution does not depend on its size.

We use the individual $a_k$ to monitor local performance, paying particular attention to the worst-performing subgroup. Let $a_1^\text{new},\dots,a_N^\text{new}$ denote the subpopulation means based on the given model's performance, and let $a_1^\text{base},\dots,a_N^\text{base}$ denote the subpopulation means based on the base model's performance. We define the (normalised) global change in performance as 
$$
\Delta G = \frac{\sum_k \left( a_k^\text{new}- a_k^{base} \right)}{\sum_k a_k^{base}}.
$$
Let $a_p^\text{new}$ be the subpopulation mean for the given classifier over its worst-performing subpopulation, and let $a_q^\text{base}$ be the subpopulation mean for the baseline classifier over its worst-performing subpopulation. Note the worst performing subpopulations can be different in the `base' and `new' cases. We define the (normalised) local change in performance as 
$$
\Delta L = \frac{a_p^\text{new}- a_q^{base}}{a_q^{base}}.
$$

We form a simple, summary measure of the relative equity of a given model with respect to a baseline model from the mean of these two differentials. Since quantification of equity of performance is the aim of this index, we only invoke it when $\Delta L > 0$, for in the absence of any local improvement the global effects are moot. We refer to this as the Holistic Equity Index (HEI):
\begin{equation}
\label{hei_equation}
HEI = \frac{ \Delta G + \Delta L }{2}.
\end{equation}

The HEI indexes the impact on the lowest performing subpopulation while taking into account the global cost across all subpopulations.
Whenever we tabulate HEI in \S\ref{demographically_fair_predictions} the base model uses empirical risk minimisation (ERM); see \S\ref{related_work}.

\section{Experiments and results}

\subsection{The data}\label{the_data}

The UK Biobank \citep{biobank} biomedical database contains a variety of brain magnetic resonance imaging (MRI) plus metadata (age, sex, etc.) from UK resident volunteers. From the T1-weighted brain imaging we randomly selected 30K unique participants (ratio of men to women 54:46, mean age 52.7 years, standard deviation 7.5 years, range 38-80 years). We shuffled and then split the data into 15K participants for training and testing CounterSynth, and a different 15K participants for the remaining down-stream tasks.

To prevent our models relying only on linear differences in head volume and shape, all volumes were
affine registered\footnote{The extent to which this necessitates affine as opposed to just rigid registration of test data depends on how much spatial augmentation is applied during training; see \S\ref{CounterSynth_training_details}. Affine registration is not a \emph{necessary} part of our method: it simply discourages CounterSynth from learning simplistic deformations. Alternatively this can be achieved through more aggressive spatial augmentations during training, but naturally this slows convergence.} to MNI152 standard space \citep{mni_space} using SPM \citep{john_unified_segmentation} and then cropped and down-sampled to $128\times 128 \times 128$ resolution. For the experiments in \S\ref{demographically_fair_predictions} we down-sampled the imaging further to $64\times64\times64$ to facilitate multiple training runs.

For the counterfactual synthesis task (\S \ref{gans_experiments}) we model age and sex, both of which are self-reported.
They can be predicted with very high accuracy from brain imaging \citep{sfcn}. For the predictive tasks (\S \ref{demographically_fair_predictions}) we use age, sex and the total volume of white matter hyperintensities (WMH). The WMH data was derived automatically by using Bianca \citep{bianca} with subsequent quality control by the UK Biobank team \citep{biobank_qc}.

Wherever we use discrete domain labels, we bin age into `younger' (age $\in[0,50]$), `middle-aged' (age $\in(51, 55]$) and `older' (age $\in(55,80]$) (these intervals are chosen to be roughly equal in size) and WMH volume into top quartile versus bottom three quartiles. Biobank's sex variable is binary.

\subsection{CounterSynth training details}\label{CounterSynth_training_details}
Our deformation generator is based on the U-Net \citep{unet} architecture adapted by \citet{dalca_voxelmorph} for diffeomorphic registration; see figure \ref{fig:total_gan}. For our discriminators we used the fully-convolutional model \citep{patch_gan}, with the 2D convolutions replaced with 3D convolutions; see figure \ref{fig:total_gan}. To train the domain classifier $D_{c l s}$ we use the `younger', `middle-aged' and `older' bins defined in \S\ref{the_data}, as well as Biobank's self-reported (binary) sex metadata.

For our experiments we shuffle the data then divide it into 80:10:10 training, validation and test splits. We train each model for 300 epochs, after which the model with the best performance on the validation set is selected. All tabulated metrics are computed on the test set. 

The batch size was 128. We used the Adam optimiser \citep{adam} with learning rate $10^{-3}$ for the generator and $2\times10^{-4}$ for the discriminator (determined based on prior experience). $L^2$ weight regularisation was applied to all the non-bias parameters, with coefficient $10^{-4}$. The models were trained on an 8-card P100-based NVIDIA DGX-1.

We used stochastic discriminator augmentation (SDA) \citep{gans_small_data}, which improves GAN performance in the absence of overwhelming amounts of training data. The augmentation functions were imported from MONAI v0.6 \citep{MONAI} and applied to each training example independently with a probability of $0.8$ (as recommended in \citet{gans_small_data}). We used random affine and elastic deformations, nonlinear histogram transformations, contrast changes and additive Gaussian noise.

In the original StarGAN framework target domain labels are sampled randomly when minimising loss \eqref{learn_counterfactuals}. In our case we have relatively few domains, so we are able to instead compute equation \eqref{learn_counterfactuals} for all possible labels, using gradient accumulation to overcome memory limitations. This noticeably improves stability without a substantial increase in training time.

\subsection{Predictive model training details}\label{predictive_model_details}
For the predictive tasks we use the official implementation of the current state of the art age and sex prediction model \citep{sfcn}, which is implemented in PyTorch \citep{pytorch}. We used the Adam \citep{adam} optimiser, with default settings, and a batch size of 128. We use the training data augmentation functions listed in \S\ref{CounterSynth_training_details}, but with a lower probability of $0.2$ (outside the SDA framework high probabilities are not required).

All models are trained five times, for 200 epochs, with different 80:7:13 training, validation and test splits (we increased the size of the test set at the expense of the validation set until it reached 2000 participants, to boost the numbers of under-represented demographics). To reiterate, the data used here does not overlap with the data used to train CounterSynth. All models are trained with oversampling of the minority classes and demographics. The model with the highest balanced accuracy on the validation set is then evaluated on the test set. The models were trained on an 8-card P100-based NVIDIA DGX-1.

Our experiments with DRO are based on empirical worst-group risk optimisation, see \S\ref{related_work}. The most effective version of this method requires large amounts of data set-specific $L^2$ regularisation (see \S3.2 of \citet{dro_neural_nets}). To find these values we performed a cross-validated grid-search using the training and validation sets. An $L^2$ regularisation coefficient of $0.01$ provided the best results for all tasks, a finding which is consistent with the values presented in \citet{dro_neural_nets}.

\subsection{Counterfactual synthesis}\label{gans_experiments}
Counterfactual synthesis cannot be evaluated by a simple image comparison, for the synthetic image is definitionally inexistent. Instead we quantify the fidelity of the conditioning biological signals --- here age and sex --- in three complementary ways. First, we use voxel-based brain morphometry (VBM) \citep{vbm} to compare their regional correlates across real and synthetic images. Second, we use a discriminative model trained exclusively on real data to compare their relative predictability. The former provides an index of the spatial fidelity of the counterfactual anatomy, the latter of its predictability from real data. Third, in the absence of likelihoods we use the Fr\'{e}chet inception distance (FID) \citep{fid}, the current standard for quantifying the overall quality of GAN-generated images.

\subsubsection{Experiment: voxel-based morphometry}\label{vbm_experiment}
VBM is conventionally used to infer the population-level anatomical correlates of a set of biological factors of interest \citep{vbm_intro, john_vbm}. This is done via a mass-univariate voxel-wise comparison of tissue concentrations across homologous regions, enabled by prior non-linear registration to a commmon stereotactic space. Here we implemented this within SPM's well-established pre-processing and statistical framework.

We used SPM's unified tissue segmentation and normalisation algorithm \citep{john_unified_segmentation} to generate non-linearly registered grey matter segmentations of 1000 real and 1000 counterfactual images conditioned on age or sex, all drawn from the test set. At each voxel, grey matter concentration, the dependent variable, was entered into a multiple regression with age, sex, origin, and total intracranial volume as independent variables. After model estimation, two one-tailed t-tests were performed on the regression coefficients (slopes) of the age and sex variables, with the resulting SPMs thresholded at $p<0.05$ FWE (cluster-based family-wise correction), and $p<0.01$ (uncorrected cluster forming threshold). An unusually lenient uncorrected threshold was deliberately chosen to reveal inferred areas to their maximum extent. Two-tailed t-tests were performed on the coefficients of the origin variable --- real or counterfactual --- separately for the age and sex contrasts, with identical thresholding. Anatomical labels based on the AAL3 atlas \citep{aal3} were assigned to the peak of each cluster, and the top 10 regions were compared.

Inspection of the resultant maps (Figure 2) shows similar anatomical patterns for all contrasts. For both age and sex, the CounterSynth VBM t-statistics matched 95\% of the anatomical labels identified in the real data. Few regions in the counterfactual vs real comparison survived the extremely lenient uncorrected threshold.

Note the fidelity quantified here is of the conditioning, background signal, not the foreground signal we seek to preserve. This is quantified by the downstream discriminative model (see \S \ref{demographically_fair_predictions}).

\begin{figure*}
\hspace*{-1cm}
    \raggedleft
    \includegraphics[width=18.5cm]{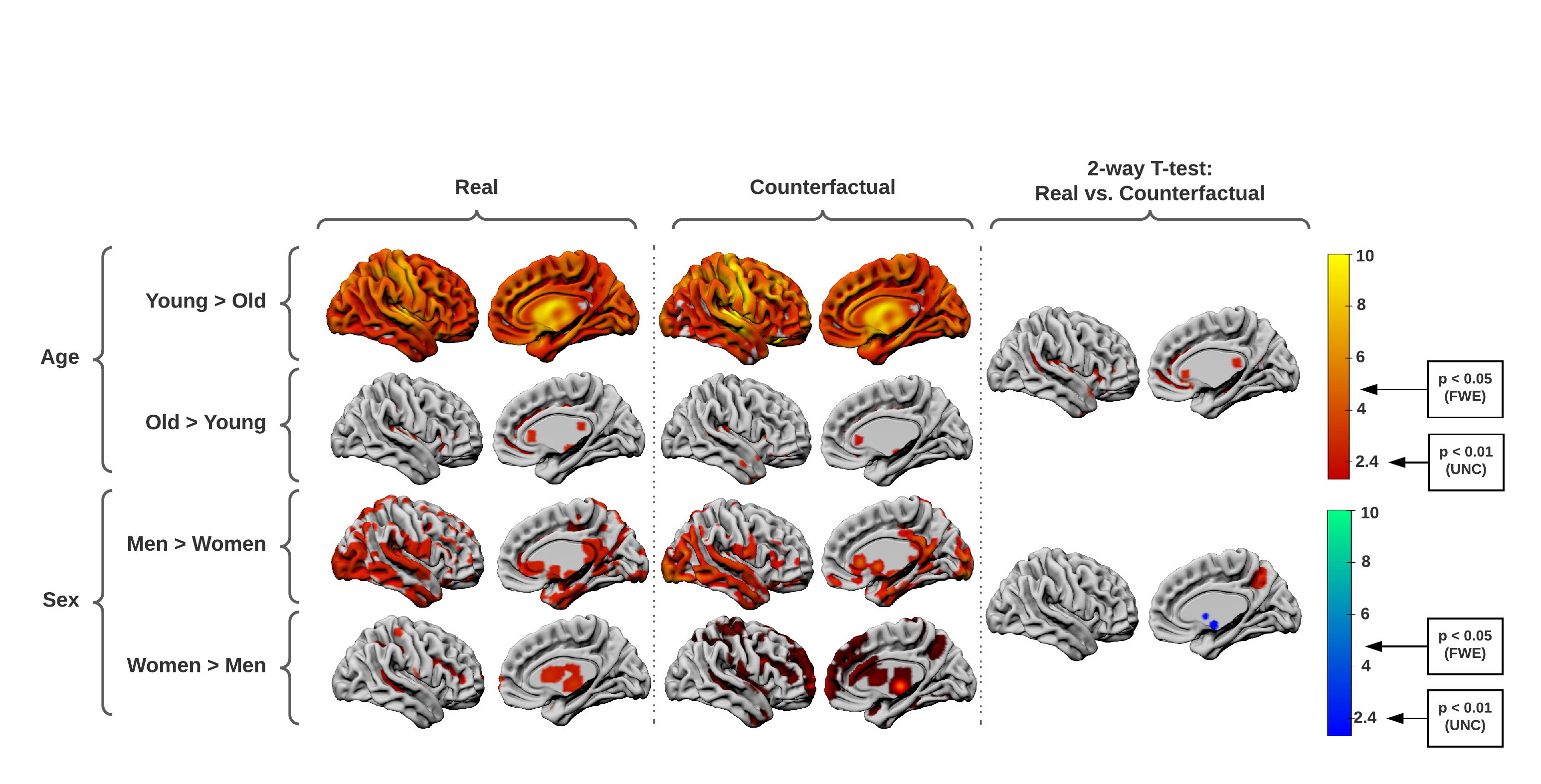}
    \caption[T-statistics maps of CounterSynth sex counterfactuals]{SPM's VBM t-statistics for grey matter changes induced by the CounterSynth age and sex deformations. Leftmost, the grey matter changes associated with age and sex in the original data. Middle, those same changes but in the synthesised counterfactuals. Rightmost, the two one-tailed post-loc t-tests that show voxels where the the real and counterfactual regression coefficients differ, the differences being negligible. There are two T-value thresholds to consider, the uncorrected estimation threshold for $p<0.01$ (UNC) and the family-wise estimation threshold for  $p<0.05$ (FWE). }
    \label{fig:spm_sex_blobs}
\end{figure*}

\begin{figure*}
\hspace*{-1cm}
    \centering
    \includegraphics[width=19cm]{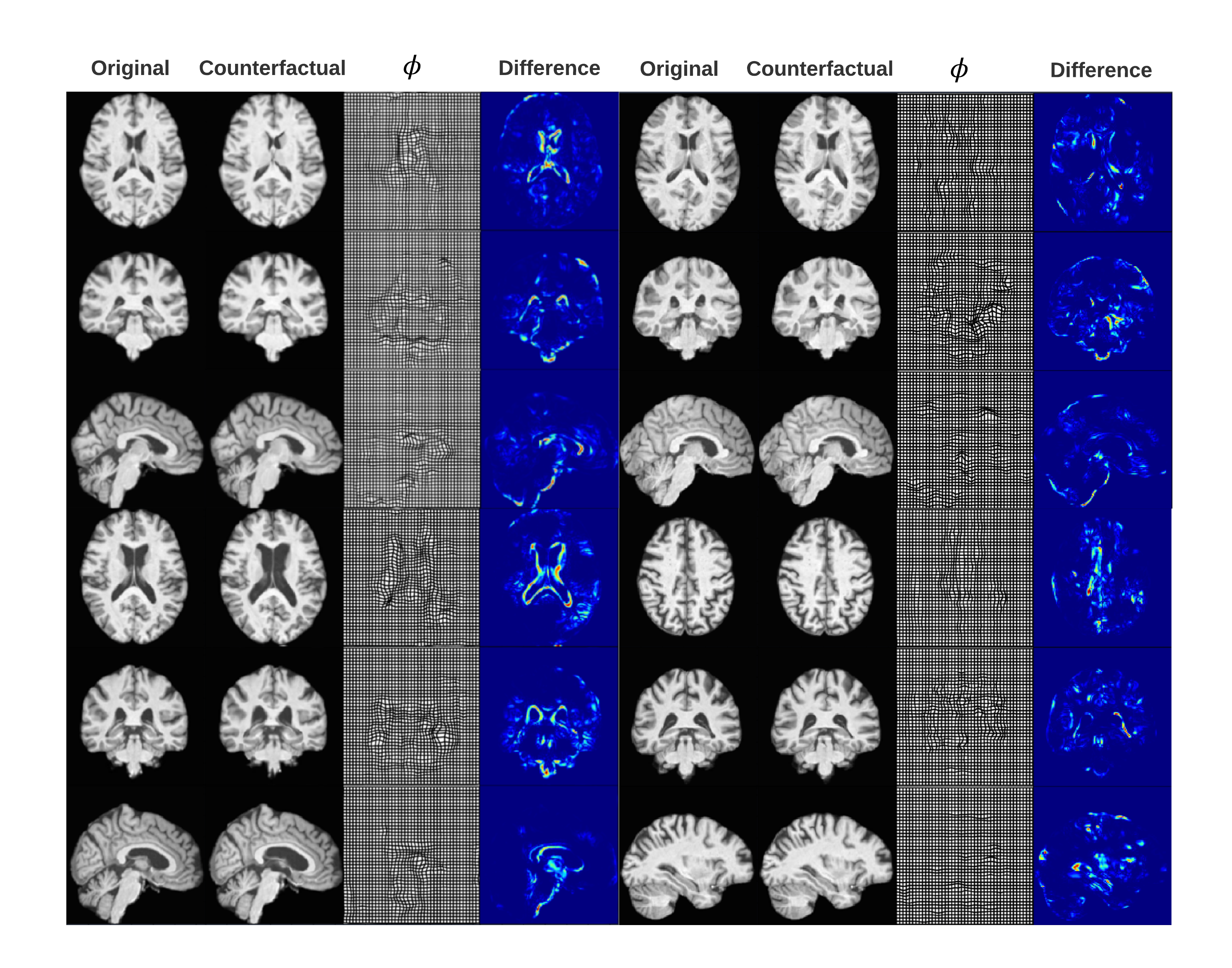}
    \caption[Age transforms]{Examples of the synthesis of volumetric counterfactuals. \textbf{Age}: The first four columns (from left to right) are age counterfactuals. The first three rows show a `middle-aged' brain and its `younger' counterfactual. The last three rows show the `older' counterfactual for a `middle-aged' brain; \textbf{Sex}: Columns five to eight show sex counterfactuals. The first three rows show a `female' brain and its `male' counterfactual. The second three rows show the `female' counterfactual for a `male' brain.}
    \label{fig:age_transforms}
\end{figure*}

\subsubsection{Experiment: age and sex prediction}\label{classification_validation}
To quantify the preservation of biological signal in the counterfactuals, we trained age and sex classifiers, and an age regression model, to measure the preservation of age and sex correlates. The models were trained once using the training participants set aside for the predictive modelling (see \S\ref{the_data}), attaining a test set accuracy of 99.2\% for sex and 99.5\% for age, and a MAE of 2.42 for age (very close to the MAE reported in \citet{sfcn}). We then use CounterSynth to create the following three sets of counterfactual examples from the test set:
\begin{enumerate}
    \item We replace each brain with its sex counterfactual.
    \item We replace each brain with its two age counterfactuals.
    \item We apply CounterSynth twice in succession: first replacing each brain with its sex counterfactual, then replacing this counterfactual with its two age counterfactuals.
\end{enumerate}

\begin{figure*}
    \centering
    \includegraphics[scale=0.5]{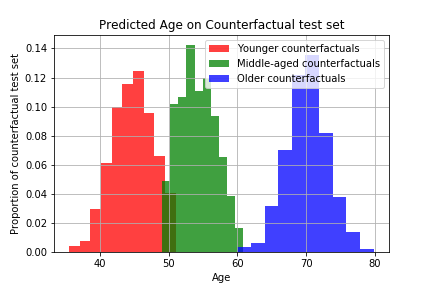}
    \caption{The distribution of predicted ages for the CounterSynth synthesised counterfactuals. {\bf Red:} Distribution of predicted ages for the `middle-aged' \& `older' participants transformed into `younger' participants; {\bf Green:} Distribution of predicted ages for the `younger' \& `older' participants transformed into `middle-aged' participants; {\bf Blue:} Distribution of predicted ages for the `younger' \& `middle-aged' participants transformed into `older' participants.}
    \label{continuous_age_histogram}
\end{figure*}

We classified the age and sex of these synthetic brains with a very similar average accuracy to those attained on the original data (The results are presented in Table \ref{classification_regression_test}). Together with the VBM maps presented in Figure \ref{fig:spm_sex_blobs} this indicates that CounterSynth is preserving significant amounts of biological signal. To give a sense of the spread of counterfactual ages, we predicted the continuous ages of the counterfactuals in set (2) above. The histogram of predictions is shown in Figure \ref{continuous_age_histogram}.

\begin{table}[h]
\centering
\begin{tabular}{lllll} \toprule
 Target & \multicolumn{1}{p{1cm}}{\centering Original \\ accuracy} & \multicolumn{1}{p{1.2cm}}{\centering Age c.f. \\ accuracy} & \multicolumn{1}{p{1.2cm}}{\centering Sex c.f. \\ accuracy} & \multicolumn{1}{p{2.2cm}}{\centering Age \& sex c.f. \\ accuracy} \\ \midrule
 Sex &  \hspace{0.05cm} 99.2\% & \hspace{0.05cm} 97.6\% & \hspace{0.1cm} 96.8\% & \hspace{0.5cm} 96.4\% \\
 Age &  \hspace{0.05cm} 99.5\% & \hspace{0.05cm} 97.7\% & \hspace{0.1cm} 97.9\% & \hspace{0.5cm} 97.5\% \\ \bottomrule
\end{tabular}
\captionof{table}{Average classification accuracy for original test set data and for CounterSynth counterfactual (c.f.) images.}
\label{classification_regression_test}
\end{table}

\subsubsection{Experiment: Fr\'{e}chet inception distance}\label{FIDS}
To provide a point of reference with other likelihood-free generative models, we compute the Fr\'{e}chet inception distance (FID) \citep{fid} between our original data and the synthesised counterfactuals.

The FID is computed using the hidden activations produced when these two sets of data are passed through an image model. We use the official FID implementation provided in \citep{fid}, which we adapted to PyTorch. This is based on the widely-used Inception v3 model trained on Imagenet. In terms of the sample means $\boldsymbol{\mu}_{orig.}$, $\boldsymbol{\mu}_{gen.}$ and covariances $\boldsymbol{\Sigma}_{orig.}$, $\boldsymbol{\Sigma}_{gen.}$ of these sets of hidden activations, and in terms of the $L^2$ norm $||\cdot||_2$ and trace operator $\text{tr}$, the FID is defined as
\begin{equation}
    ||\boldsymbol{\mu}_{orig.} - \boldsymbol{\mu}_{gen.}||_2^2 + \text{tr}(\boldsymbol{\Sigma}_{orig.} + \boldsymbol{\Sigma}_{gen.} - 2\sqrt{\boldsymbol{\Sigma}_{orig.}\boldsymbol{\Sigma}_{gen.}}).
\end{equation}
This FID applies only to 2D images, so we extracted a $128\times128$ slice along each axis in turn. We chose the slice with the maximum voxel-wise t-statistic for the relevant attribute (see \S\ref{vbm_experiment}). The results are presented in Table \ref{fid_table}, together with the FIDs of the 3D-StyleGAN (\citet{3d_style_gan}; see also \S\ref{volumetric_modelling_with_GANs}) for comparison.

\begin{table}[h]
\centering
\begin{tabular}{llll} \toprule
 Model & Axial & Coronal & Sagittal \\ \midrule
 CounterSynth (age) &  11.9 & 10.7 & 9.5 \\
 CounterSynth (sex) &  9.8 & 9.9 & 9.1 \\
 CounterSynth (age \& sex) &  12.4 & 11.1 & 9.8 \\
 3D-StyleGAN &  71.3 & 88.5 & 106.9 \\ \bottomrule
\end{tabular}
\caption{Fr\'{e}chet inception distances between original data and synthetic data. Lower values correspond to greater visual (metrical and perceptual) similarity. CounterSynth's very low FIDs are almost inevitable because regularised deformations leave much of the original image essentially unchanged, which is part of our motivation for using them.
}\label{fid_table}
\end{table}

\subsection{Downstream predictive equity}\label{demographically_fair_predictions}
In this section, we demonstrate CounterSynth's ability to improve the average performance, and the worst-demographic-subpopulation performance, of a classifier trained on demographically-imbalanced data. We also demonstrate CounterSynth's ability to lessen the extent to which a classifier learns spurious correlations between demographic attributes and the target label.

We compare the following approaches based on ERM, group DRO (\S\ref{group_dro_section}), confounder-free networks (\S\ref{cf_net_section}) and CounterSynth training augmentation (\S\ref{methodology}). To avoid information leaks, we use two CounterSynth models, one trained exclusively to produce age counterfactuals, and one to produce sex counterfactuals. This ensures that the age counterfactuals do not carry over any sex information and vice-versa. 
In all cases, where necessary we oversample each sex and each age group until equal numbers are attained. We also perform minority class oversampling to balance the representation of the classes being predicted.

\begin{itemize}
    \item \textbf{ERM}: In line with empirical risk minimisation (see \S\ref{related_work}), we simply use online stochastic gradient descent (SGD) to optimise the predictive models.
    \item \textbf{DRO}: We use the official implementation of the group DRO optimiser (Algorithm 1 in \cite{dro_neural_nets}, which has a similar run time to SGD) with an $L^2$ regularisation coefficient determined by grid-searching (\S3.2 of \cite{dro_neural_nets}). In the group DRO setting one selects demographic attributes believed to be spuriously correlated with the target variable; see \S2.1 of \citep{dro_neural_nets}. In the case of sex classification (\S\ref{preds_class}) we choose age, in the case of WMH volume classification (\S\ref{confounder_predictions}-\S\ref{collider_bias}) we choose age and sex.
    \item \textbf{ERM with balancing counterfactuals (ERM-BCF)}: We use the ERM approach above, but we augment the training (validation) set with counterfactuals synthesised by CounterSynth from the training (validation) set until the under-represented demographic is as numerous as the others. A new set of counterfactuals is synthesised each epoch from equal numbers of the not-under-represented demographics.
    \item \textbf{ERM with all counterfactuals (ERM-ACF)}: We use the ERM-BCF approach, but we augment with \emph{all possible} age counterfactuals when rectifying an age imbalance and \emph{all possible} sex counterfactuals when rectifying an imbalance in the sexes. The data remains imbalanced, so we then oversample to equalise the counts of each sex and age subpopulation.
    \item \textbf{DRO-BCF}, \textbf{DRO-ACF}: As above, but with the ERM approach replaced with the DRO approach.
    \item \textbf{CF-Net}: We use the official implementation of the Confounder-Free network (CF-Net), \citep{confounder_free}; see \S\ref{cf_net_section}. CF-Net learns a featurisation of the data that is approximately invariant to a chosen demographic attribute. In the case of sex classification (\S\ref{preds_class}) and WMH volume classification (\S\ref{confounder_predictions}) we choose age. The official implementation does not support regression objectives (\S \ref{age_prediction}) or tasks with multiple confounders (\S \ref{collider_bias}), so we omit comparisons with CF-Net in these cases. CF-Net also takes significantly longer to converge than the other baseline models: we therefore leave training it in conjunction with BCF and ACF to the future.
\end{itemize}

\subsubsection{Experiment: sex classification}\label{preds_class}\label{sex_prediction}
For this experiment we create training sets with different levels of missingness of a chosen demographic (older people), and use them to train a classifier to predict sex.
We use the seven approaches described in the previous section to counter the negative effects of the resultant demographic imbalance on average performance, and to boost worst-demographic-subpopulation performance, as measured by (balanced) accuracy, precision and recall.

To simulate the missingness we create sets with the maximum possible equal number of `younger' and `middle-aged' participants (see \S\ref{the_data} for age ranges), then add `older' participants until they constitute a given percentage of the total. The percentages are 0, 1, 10 and 25. There are equal numbers of men and women in the `younger', 'middle-aged' and `older' subpopulations.

For each combination of model and missingness we present in Table \ref{sex_classification_table} (with standard deviation) balanced accuracy, precision and recall for the best- and worst-preforming demographics, average balanced accuracy, and our HEI index \eqref{hei_equation}.

\begin{table*}[!t]
\centering
\begin{tabular}{lllllll} \toprule
 Method & Best B-Acc & Best Prec/Recall & Worst B-Acc & Worst Prec/Recall & Avg. B-Acc & HEI \\ \midrule
 
 ERM, 0\% & 84.0 $\pm$ 1.1 & 90.3/84.6 $\pm$ 2.5/2.5 & 76.0 $\pm$ 1.0 & 73.0/72.7 $\pm$ 3.9/2.9 & 77.5 $\pm$ 1.1 & N/A \\
 
 DRO, 0\% & 86.8 $\pm$ 1.4 & 91.7/84.3 $\pm$ 1.6/0.9 & 76.8 $\pm$ 1.4 & 76.0/75.2 $\pm$ 1.5/1.1 & 79.5 $\pm$ 1.0 & 1.8 \\
 
 CF-Net, 0\% & 84.9 $\pm$ 1.8 & 92.1/81.3 $\pm$ 1.1/1.3 & 76.1 $\pm$ 1.1 & 77.3/72.2 $\pm$ 1.8/1.2 & 78.6 $\pm$ 0.6 & 0.4 \\
 
 ERM-ACF, 0\% & \textbf{91.8} $\pm$ 2.2 & 93.8/\textbf{87.9} $\pm$ 0.9/4.5 & \textbf{83.9} $\pm$ 1.9 & 84.5/\textbf{83.1} $\pm$ 1.1/2.6 & \textbf{86.4} $\pm$ 1.5 & \textbf{10.9} \\
 
 DRO-ACF, 0\% & 86.9 $\pm$ 2.5 & \textbf{94.5}/83.1 $\pm$ 1.3/2.3 & 82.4 $\pm$ 2.2 & \textbf{85.9}/73.3 $\pm$ 1.5/4.0 & 83.8 $\pm$ 1.9 & 8.3 \\ \midrule

 ERM, 1\% & 88.3 $\pm$ 2.6 & 96.9/88.2 $\pm$ 1.2/3.1 & 78.6 $\pm$ 1.2 & 78.4/71.1 $\pm$ 5.0/8.0 & 81.3 $\pm$ 0.3 & N/A \\
 
 DRO, 1\% & 88.4 $\pm$ 5.3 & 93.0/88.3 $\pm$ 2.1/1.4 & 80.3 $\pm$ 3.3 & 77.4/77.4 $\pm$ 6.5/4.4 & 81.9 $\pm$ 4.0 & 1.5 \\
 
 CF-Net, 1\% & 86.3 $\pm$ 3.3 & 96.3/85.1 $\pm$ 1.9/2.3 & 79.8 $\pm$ 2.1 & 79.8/70.2 $\pm$ 3.2/3.8 & 81.4 $\pm$ 4.4 & 0.5 \\
 
 ERM-ACF, 1\% & \textbf{92.9} $\pm$ 2.2 & \textbf{98.2}/90.2 $\pm$ 2.6/5.4 & 85.5 $\pm$ 1.1 & 85.5/79.9 $\pm$ 2.7/6.2 & 87.3 $\pm$ 0.5 & 8.1 \\
 
 DRO-ACF, 1\% & 89.2 $\pm$ 4.2 & 97.2/84.9 $\pm$ 3.9/0.3 & 83.5 $\pm$ 2.1 & 84.7/73.6 $\pm$ 3.2/5.3 & 85.1 $\pm$ 0.2 & 5.5 \\ 
 
 ERM-BCF, 1\% & 91.8 $\pm$ 3.9 & 89.2/\textbf{93.3} $\pm$ 4.2/3.7 & 88.9 $\pm$ 3.3 & 87.6/\textbf{90.3} $\pm$ 3.5/4.8 & \textbf{90.2} $\pm$ 3.6 & \textbf{12.7} \\
 
 DRO-BCF, 1\% & 91.5 $\pm$ 3.8 & 92.8/88.4 $\pm$ 4.8/3.8 & \textbf{89.4} $\pm$ 3.9 & \textbf{89.9}/85.5 $\pm$ 3.8/4.1 & 89.7 $\pm$ 3.3 & \textbf{12.7} \\ \midrule

 ERM, 10\% & 86.0 $\pm$ 6.8 & 94.8/87.6 $\pm$ 1.9/5.4 & 78.7 $\pm$ 4.4 & 76.8/72.3 $\pm$ 2.9/5.2 & 80.2 $\pm$ 4.1 & N/A \\
 
 DRO, 10\% & 87.4 $\pm$ 3.4 & 90.9/80.8 $\pm$ 2.4/5.2 & 83.0 $\pm$ 2.8 & 84.0/75.9 $\pm$ 2.6/5.4 & 84.4 $\pm$ 2.8 & 5.4 \\
 
 CF-Net, 10\% & 86.3 $\pm$ 2.9 & 95.2/78.9 $\pm$ 1.8/5.5 & 81.3 $\pm$ 3.3 & 82.9/70.7 $\pm$ 2.2/5.1 & 83.2 $\pm$ 3.4 & 2.9 \\
 
 ERM-ACF, 10\% & 94.0 $\pm$ 1.5 & 92.8/93.5 $\pm$ 5.9/4.3 & 86.9 $\pm$ 2.4 & 85.5/85.1 $\pm$ 3.0/3.7 & 88.7 $\pm$ 1.5 & 10.5 \\
 
 DRO-ACF, 10\% & 90.0 $\pm$ 2.8 & 97.1/86.1 $\pm$ 3.3/2.2 & 84.8 $\pm$ 2.1 & 88.2/75.2 $\pm$ 2.9/4.7 & 87.2 $\pm$ 0.6 & 8.2 \\
 
 ERM-BCF, 10\% & 94.5 $\pm$ 3.1 & 92.0/\textbf{95.2} $\pm$ 3.4/3.2 & \textbf{91.3} $\pm$ 3.1 & 90.3/\textbf{92.9} $\pm$ 4.4/3.2 & \textbf{92.1} $\pm$ 3.1 & \textbf{14.4} \\
 
 DRO-BCF, 10\% & \textbf{95.5} $\pm$ 1.3 & \textbf{98.1}/92.3 $\pm$ 1.8/2.7 & \textbf{91.3} $\pm$ 1.0 & \textbf{92.6}/87.5 $\pm$ 2.9/0.0 & 91.2 $\pm$ 0.5 & 14.1 \\ \midrule

 ERM, 25\% & 93.5 $\pm$ 2.2 & 91.9/93.1 $\pm$ 5.7/3.7 & 85.2 $\pm$ 4.0 & 83.4/84.4 $\pm$ 3.7/3.7 & 87.4 $\pm$ 2.7 & N/A \\
 
 DRO, 25\% & 88.9 $\pm$ 3.6 & 88.0/94.0 $\pm$ 1.6/1.7 & 86.3 $\pm$ 1.9 & 80.7/86.8 $\pm$ 5.7/1.6 & 87.6 $\pm$ 2.3 & 0.8 \\
 
 CF-Net, 25\% & 87.3 $\pm$ 3.1 & 91.2/89.4 $\pm$ 1.9/2.3 & 85.8 $\pm$ 2.4 & 84.7/81.3 $\pm$ 4.3/3.5 & 87.6 $\pm$ 2.1 & 0.4 \\
 
 ERM-ACF, 25\% & 96.0 $\pm$ 1.4 & 92.5/\textbf{97.8} $\pm$ 4.5/1.7 & 90.2 $\pm$ 1.2 & 88.0/88.7 $\pm$ 2.1/1.9 & 92.0 $\pm$ 1.3 & 5.6 \\
 
 DRO-ACF, 25\% & 91.9 $\pm$ 1.2 & 95.7/86.6 $\pm$ 0.9/2.6 & 88.0 $\pm$ 2.1 & 90.8/81.8 $\pm$ 3.9/2.7 & 89.2 $\pm$ 1.8 & 2.7 \\
 
 ERM-BCF, 25\% & 95.3 $\pm$ 4.5 & 92.0/97.3 $\pm$ 3.8/1.3 & \textbf{91.1} $\pm$ 2.9 & 89.0/\textbf{93.5} $\pm$ 3.0/2.9 & \textbf{92.4} $\pm$ 3.3 & \textbf{10.5} \\
 
 DRO-BCF, 25\% & \textbf{96.9} $\pm$ 3.2 & \textbf{96.5}/96.4 $\pm$ 5.0/2.5 & 89.9 $\pm$ 4.4 & \textbf{93.4}/86.7 $\pm$ 2.2/3.8 & 91.9 $\pm$ 3.2 & 9.5 \\ \bottomrule
 
\end{tabular}
\caption{Test set results for sex classification with varying representations of `older' participants. The number following the method indicates the percentage of `older' participants retained in the training and validation sets (0, 1, 10 or 25\%).  The number of `young' and `middle-aged' patients in the training and validation sets is 5153, 452 respectively. Of the `older' participants in the training and validation sets respectively, 1\% amounts to 52, 4 participants, 10\% amounts to 572, 50 participants, and 25\% amounts to 1717, 151 participants. Here `N/A' indicates that $\Delta L\leq0$ (see \S\ref{cost_index}), so the HEI does not apply.}\label{sex_classification_table}
\end{table*}

Table \ref{sex_classification_table} shows that training with synthesised counterfactuals improves performance in every case. As the representation of the `older' subpopulation increases, DRO outperforms ERM, but at 0\% and 1\% representations the improvements are marginal. The worst-performing demographic significantly improves whenever counterfactuals are added, without compromising the best-performing demographic, or average performance. The `BCF' counterfactual strategy leads to the greatest improvements to the worst-performing demographic and to the average population. In all setups, CF-Net is better than ERM, but worse than all other methods.
Both counterfactual augmentation strategies consistently improve the performance of ERM and DRO.

\subsubsection{Experiment: age regression}\label{age_prediction}
For this experiment we create training sets with different sex imbalances, and use them to train an age prediction model.
We test the approaches described at the start of \S\ref{demographically_fair_predictions} to counter the negative effect on average performance and to boost worst-demographic-subpopulation performance, as measured by mean absolute error (MAE).
We do not use group DRO because the official algorithm was unstable in conjunction with our regression objectives, and we do not use CF-Net because the official implementation does not support a regression objective. To simulate the missingness we create sets with the maximum possible number of female participants and equal numbers of `younger', `middle-aged' and `older' participants; see \S\ref{the_data}. Then we add male participants until they constitute a given percentage of the total. The percentages are 0, 1, 10 and 25.

\begin{table*}[!t]
\centering
\begin{tabular}{llll} \toprule
 Method & Men MAE & Women MAE \\ \midrule
 
 ERM, 0\% & 5.14 $\pm$ 0.7 & 3.15 $\pm$ 0.44\\
 \textbf{ERM-ACF}, 0\% & \textbf{3.89} $\pm$ 0.40 & \textbf{3.13} $\pm$ 0.09\\ \midrule
 
 ERM, 1\% & 4.27 $\pm$ 0.31 & 3.34 $\pm$ 0.25 \\
 ERM-ACF, 1\% & 3.67 $\pm$ 0.28 & 3.23 $\pm$ 0.15 \\ 
 \textbf{ERM-BCF}, 1\% & \textbf{3.51} $\pm$ 0.25 & \textbf{3.03} $\pm$ 0.04 \\ \midrule
 
 ERM, 10\% & 3.92 $\pm$ 0.35 & 3.01 $\pm$ 0.33 \\
 ERM-ACF, 10\% & 3.44 $\pm$ 0.16 & 2.92 $\pm$ 0.11 \\ 
 \textbf{ERM-BCF}, 10\% & \textbf{3.43} $\pm$ 0.31 & \textbf{2.91} $\pm$ 0.05 \\ \midrule
 
 ERM, 25\% & 3.80 $\pm$ 0.26 & 2.93 $\pm$ 0.04 \\
 ERM-ACF, 25\% & 3.20 $\pm$ 0.23 & 2.91 $\pm$ 0.09 \\ 
 \textbf{ERM-BCF}, 25\% & \textbf{2.92} $\pm$ 0.11 & \textbf{2.89} $\pm$ 0.12 \\ \bottomrule
 
\end{tabular}
\captionof{table}{Mean absolute errors over the test set for age regression with varying representation of male participants. The number of women in the training and validation sets is 7627, 846 respectively. The number of men in the training and validation sets respectively for the different percentages are, for 2.5\%, 195, 21; for 5\%, 404, 44; for 10\%, 846 and 94; for 25\%, 2542, 282.}
\label{age_regression_table}
\end{table*}

Table \ref{age_regression_table} shows that counterfactual augmentation drastically reduces the model's error rate on the under-represented subpopulation, while also consistently improving its performance on the rest of the population.

\subsubsection{Experiment: confounders}
\label{confounder_predictions}

In \S\ref{sex_prediction} and \S\ref{age_prediction} we demonstrated CounterSynth's ability to rectify poor population and worst-subpopulation performance given a demographically imbalanced training set. In those experiments the demographic attribute was not correlated with the target label - in this section we examine what happens when it is.

A correlation between demographic and pathological labels is common in medical imaging. Many neurological disorders, such as neurovascular and neurodegenerative disorders, exhibit marked correlation with age \citep{small_vessel_age, alzheimers_age}; others, such as neuroinflammatory disorders, with sex \citep{sex_diff_neuroinflammation}. A good example from UK Biobank imaging data is WMH volume: older participants tend to have higher WMH volumes (sample Pearson correlation coefficient of $0.38$ with $p < 0.0005$). Studies  have highlighted possible associations between abnormally high WMH volumes and risks of stroke, cognitive decline and dementia \citep{wmh_1, wmh_2}.

In this section we simulate various age imbalances while training a classifier to predict whether a participant's WHM volume is in the bottom three quartiles of the population versus the top quartile (see \S \ref{the_data}). We test the six approaches described at the start of \S\ref{demographically_fair_predictions} to counter the negative effects of this imbalance on performance and equity. In the first experiment the population defined by the demographic attribute most strongly correlated with the target, the `older' participants, is under-represented in the training and validation data. The population defined by the demographic attribute least correlated with the target, the `younger' participants, is over-represented. In the second experiment the converse is true. In both experiments we vary the ratio of both demographics. The results are presented in Tables \ref{abnormal_wmh_varying_older} and \ref{abnormal_wmh_varying_younger}.

\begin{table*}[!t]
\centering
\begin{tabular}{lllllll} \toprule
 Method & Best B-Acc & Best Prec/Recall & Worst B-Acc & Worst Prec/Recall & Avg. B-Acc & HEI \\ \midrule
 
 ERM, 1\% & 62.9 $\pm$ 3.8 & 57.8/62.9 $\pm$ 2.5/4.8 & 57.6 $\pm$ 2.6 & 55.0/57.6 $\pm$ 2.9/2.6 & 64.8 $\pm$ 3.2 & N/A \\
 
 DRO, 1\% & 60.8 $\pm$ 2.6 & 61.0/60.8 $\pm$ 2.7/2.6 & 56.9 $\pm$ 5.4 & 55.2/56.9 $\pm$ 5.0/5.4 & 64.4 $\pm$ 2.1 & N/A \\
 
 CF-Net, 1\% & 59.9 $\pm$ 3.3 & 62.3/58.5 $\pm$ 2.2/3.1 & 56.3 $\pm$ 4.6 & 57.3/52.1 $\pm$ 5.4/4.8 & 64.1 $\pm$ 2.8 & N/A \\
 
 ERM-ACF, 1\% & \textbf{67.3} $\pm$ 2.6 & 65.3/\textbf{67.3} $\pm$ 4.6/2.6 & \textbf{65.3} $\pm$ 4.3 & \textbf{63.8}/65.3 $\pm$ 4.2/4.3 & 68.3 $\pm$ 5.1 & \textbf{8.0} \\
 
 DRO-ACF, 1\% & 65.2 $\pm$ 5.0 & 64.0/65.2 $\pm$ 3.4/5.0 & 63.7 $\pm$ 3.9 & 57.5/63.7 $\pm$ 3.8/3.9 & 67.9 $\pm$ 4.8 & 5.2 \\
 
 ERM-BCF, 1\% & 65.4 $\pm$ 2.9 & \textbf{66.1}/65.4 $\pm$ 11.5/2.9 & 64.7 $\pm$ 3.5 & 62.6/\textbf{64.7} $\pm$ 2.5/3.5 & \textbf{69.5} $\pm$ 3.3 & 6.8 \\
 
 DRO-BCF, 1\% & 64.4 $\pm$ 3.0 & 64.5/64.4 $\pm$ 3.1/3.0 & 57.7 $\pm$ 8.2 & 52.3/57.7 $\pm$ 3.7/8.2 & 68.0 $\pm$ 4.4 & 3.3 \\ \midrule
 
 ERM, 10\% & 63.6 $\pm$ 3.5 & 64.0/63.6 $\pm$ 3.9/3.5 & 59.6 $\pm$ 2.4 & 59.4/59.6 $\pm$ 5.8/2.4 & 69.1 $\pm$ 3.9 & N/A \\
 
 DRO, 10\% & 64.1 $\pm$ 3.2 & 64.1/63.2 $\pm$ 2.3/3.4 & 57.3 $\pm$ 3.1 & 53.7/57.3 $\pm$ 3.9/3.8 & 66.5 $\pm$ 4.2 & N/A \\
 
 CF-Net, 10\% & 63.9 $\pm$ 4.1 & 65.4/62.1 $\pm$ 2.8/3.1 & 56.7 $\pm$ 3.3 & 55.2/53.2 $\pm$ 4.1/3.3 & 65.2 $\pm$ 3.8 & N/A \\
 
 ERM-ACF, 10\% & \textbf{68.1} $\pm$ 3.4 & \textbf{67.5}/\textbf{68.1} $\pm$ 4.4/3.4 & \textbf{65.0} $\pm$ 6.1 & \textbf{63.2}/\textbf{65.0} $\pm$ 5.0/6.1 & 71.3 $\pm$ 1.7 & \textbf{12.2} \\
 
 DRO-ACF, 10\% & 65.0 $\pm$ 3.1 & 65.1/65.0 $\pm$ 2.9/3.1 & 58.7 $\pm$ 4.1 & 54.8/58.7 $\pm$ 4.2/4.1 & 67.9 $\pm$ 4.7 & N/A \\
 
 ERM-BCF, 10\% & 66.5 $\pm$ 6.5 & 66.3/66.5 $\pm$ 3.2/3.5 & 64.6 $\pm$ 4.8 & 62.6/64.6 $\pm$ 4.2/4.8 & 70.3 $\pm$ 2.3 & 10.8 \\
 
 DRO-BCF, 10\% & 67.2 $\pm$ 2.3 & 67.0/67.2 $\pm$ 2.3/2.3 & 62.3 $\pm$ 7.7 & 56.3/62.3 $\pm$ 3.8/7.7 & \textbf{71.4} $\pm$ 1.4 & 8.3 \\ \midrule

 ERM, 25\% & 63.5 $\pm$ 5.0 & 63.2/63.5 $\pm$ 6.1/5.0 & 59.0 $\pm$ 5.2 & 61.7/59.0 $\pm$ 7.9/5.2 & 69.1 $\pm$ 5.6 & N/A \\
 
 DRO, 25\% & 65.7 $\pm$ 4.6 & 65.2/65.7 $\pm$ 3.8/4.6 & 63.1 $\pm$ 4.0 & 60.2/63.1 $\pm$ 1.7/4.0 & 69.4 $\pm$ 3.7 & 6.0 \\
 
 CF-Net, 25\% & 64.9 $\pm$ 3.9 & 66.3/62.3 $\pm$ 4.1/4.2 & 62.7 $\pm$ 4.0 & 63.1/59.8 $\pm$ 2.3/3.4 & 69.2 $\pm$ 2.8 & 2.0 \\
 
 ERM-ACF, 25\% & 67.7 $\pm$ 3.6 & 68.7/67.7 $\pm$ 3.3/3.6 & 64.6 $\pm$ 3.9 & 65.4/64.6 $\pm$ 5.2/3.9 & \textbf{73.0} $\pm$ 3.5 & 8.5 \\
 
 DRO-ACF, 25\% & 65.5 $\pm$ 2.4 & 67.7/65.5 $\pm$ 2.4/2.4 & 60.8 $\pm$ 3.1 & 54.9/60.8 $\pm$ 2.9/6.4 & 67.7 $\pm$ 3.5 & 3.5 \\
 
 ERM-BCF, 25\% & 66.5 $\pm$ 3.5 & 64.8/66.5 $\pm$ 1.4/3.5 & 66.5 $\pm$ 3.5 & 64.8/66.5 $\pm$ 1.4/3.5 & 70.0 $\pm$ 4.2 & 13.2 \\
 
 DRO-BCF, 25\% & \textbf{69.7} $\pm$ 3.9 & \textbf{68.6}/\textbf{69.7} $\pm$ 7.1/3.9 & \textbf{67.9} $\pm$ 4.0 & \textbf{65.4}/\textbf{67.9} $\pm$ 1.7/4.0 & 71.6 $\pm$ 4.6 & \textbf{14.6} \\ \bottomrule

\end{tabular}
\caption{Test set results for WMH volume classification with varying levels of imbalance for `older' participants. The number following the method indicates the percentage of ‘older’participants retained in the training and validation sets (1, 10 or 25\%). The number of `young' and `middle' patients in the training and validation sets is 5153, 452 respectively. Of the `older' participants in the training and validation sets respectively, 1\% amounts to 52, 4 participants, 10\% amounts to 572, 50 participants, and 25\% amounts to 1717, 151 participants. Here `N/A' indicates that $\Delta L\leq0$ (see \S\ref{cost_index}), so the HEI does not apply.}\label{abnormal_wmh_varying_older}
\end{table*}

\begin{table*}[!t]
\centering
\begin{tabular}{lllllll} \toprule
 Method & Best B-Acc & Best Prec/Recall & Worst B-Acc & Worst Prec/Recall & Avg. B-Acc & HEI \\ \midrule
 
 ERM, 1\% & 69.8 $\pm$ 4.6 & 70.0/69.8 $\pm$ 5.0/4.6 & 63.1 $\pm$ 2.7 & 67.2/63.1 $\pm$ 2.0/2.7 & 66.9 $\pm$ 4.3 & N/A \\
 
 DRO, 1\% & 60.7 $\pm$ 1.1 & 59.9/60.7 $\pm$ 1.7/1.1 & 59.0 $\pm$ 3.1 & 55.2/59.0 $\pm$ 1.0/3.1 & 59.7 $\pm$ 2.5 & N/A \\
 
 CF-Net, 1\% & 62.4 $\pm$ 3.1 & 63.9/60.9 $\pm$ 3.7/2.3 & 61.0 $\pm$ 3.4 & 61.4/58.8 $\pm$ 3.2/2.8 & 61.3 $\pm$ 3.1 & N/A \\
 
 ERM-ACF, 1\% & \textbf{75.0} $\pm$ 1.3 & \textbf{75.3}/\textbf{75.0} $\pm$ 1.2/1.3 & 68.2 $\pm$ 0.6 & \textbf{71.7}/68.2 $\pm$ 1.8/0.6 & \textbf{72.1} $\pm$ 0.3 & 7.9 \\
 
 DRO-ACF, 1\% & 70.3 $\pm$ 3.8 & 67.0/70.3 $\pm$ 1.5/3.8 & 64.4 $\pm$ 3.5 & 59.5/64.4 $\pm$ 3.0/3.5 & 67.3 $\pm$ 1.0 & 1.3 \\
 
 ERM-BCF, 1\% & 73.6 $\pm$ 1.1 & 74.0/73.6 $\pm$ 1.5/1.1 & \textbf{69.7} $\pm$ 5.0 & 70.9/\textbf{69.7} $\pm$ 6.6/5.0 & 72.0 $\pm$ 1.4 & \textbf{9.0} \\
 
 DRO-BCF, 1\% & 68.4 $\pm$ 2.9 & 65.3/68.4 $\pm$ 4.0/2.9 & 64.0 $\pm$ 2.2 & 59.2/64.0 $\pm$ 2.6/2.2 & 65.9 $\pm$ 3.7 & 0 \\ \midrule

 ERM, 10\% & 70.9 $\pm$ 1.8 & 71.4/70.9 $\pm$ 1.4/1.8 & 61.8 $\pm$ 1.7 & 60.5/61.8 $\pm$ 2.7/1.7 & 67.4 $\pm$ 1.1 & N/A \\
 
 DRO, 10\% & 65.2 $\pm$ 4.7 & 64.4/65.2 $\pm$ 4.1/4.7 & 63.3 $\pm$ 2.9 & 58.3/63.3 $\pm$ 3.2/2.9 & 63.9 $\pm$ 4.2 & -1.4 \\
 
 CF-Net, 10\% & 66.4 $\pm$ 3.8 & 69.2/60.2 $\pm$ 3.8/3.2 & 63.5 $\pm$ 3.3 & 62.3/61.2 $\pm$ 4.1/3.4 & 64.2 $\pm$ 4.6 & -0.8 \\
 
 ERM-ACF, 10\% & 73.8 $\pm$ 1.5 & 73.8/73.8 $\pm$ 1.7/1.5 & \textbf{68.8} $\pm$ 3.5 & \textbf{70.8}/\textbf{68.8} $\pm$ 4.5/3.5 & \textbf{71.6} $\pm$ 1.8 & \textbf{8.8} \\
 
 DRO-ACF, 10\% & 66.9 $\pm$ 7.7 & 66.6/66.9 $\pm$ 7.5/7.7 & 61.7 $\pm$ 5.9 & 59.0/61.7 $\pm$ 5.8/5.9 & 64.4 $\pm$ 6.8 & N/A \\
 
 ERM-BCF, 10\% & \textbf{74.4} $\pm$ 4.1 & \textbf{74.7}/\textbf{74.4} $\pm$ 3.6/4.1 & 68.1 $\pm$ 1.7 & 69.9/68.1 $\pm$ 3.0/1.7 & 70.9 $\pm$ 2.8 & 7.7 \\
 
 DRO-BCF, 10\% & 70.5 $\pm$ 3.8 & 68.1/70.5 $\pm$ 2.6/3.8 & 63.8 $\pm$ 1.7 & 59.1/63.8 $\pm$ 1.3/1.7 & 67.6 $\pm$ 1.3 & 1.8 \\ \midrule

 ERM, 25\% & 70.5 $\pm$ 3.5 & 70.7/70.5 $\pm$ 3.9/3.5 & 66.5 $\pm$ 2.0 & 65.7/66.5 $\pm$ 6.6/2.0 & 68.0 $\pm$ 5.0 & N/A \\
 
 DRO, 25\% & 69.0 $\pm$ 1.4 & 69.5/69.0 $\pm$ 1.6/1.4 & 66.2 $\pm$ 2.6 & 63.8/66.2 $\pm$ 4.5/2.6 & 67.6 $\pm$ 0.1 & N/A \\
 
 CF-Net, 25\% & 69.6 $\pm$ 2.3 & 72.3/65.2 $\pm$ 2.8/2.5 & 66.7 $\pm$ 2.3 & 66.2/61.9 $\pm$ 4.1/3.8 & 67.9 $\pm$ 3.2 & 0.1 \\
 
 ERM-ACF, 25\% & \textbf{75.6} $\pm$ 1.4 & \textbf{76.3}/\textbf{75.6} $\pm$ 1.4/1.4 & 68.6 $\pm$ 3.0 & 72.6/68.6 $\pm$ 2.1/3.0 & 72.7 $\pm$ 1.2 & 5.0 \\
 
 DRO-ACF, 25\% & 69.5 $\pm$ 4.0 & 69.3/69.5 $\pm$ 3.9/4.0 & 66.7 $\pm$ 4.1 & 64.7/66.7 $\pm$ 3.0/4.1 & 68.1 $\pm$ 3.0 & 0.2 \\
 
 ERM-BCF, 25\% & 74.8 $\pm$ 1.8 & 75.5/74.8 $\pm$ 0.9/1.8 & \textbf{70.8} $\pm$ 3.0 & \textbf{73.8}/\textbf{70.8} $\pm$ 1.1/7.0 & \textbf{73.4} $\pm$ 0.8 & \textbf{7.2} \\
 
 DRO-BCF, 25\% & 70.7 $\pm$ 3.2 & 71.5/70.2 $\pm$ 3.2/3.5 & 66.9 $\pm$ 4.5 & 67.9/68.3 $\pm$ 3.2/3.7 & 69.2 $\pm$ 2.8 & 0.8 \\ \bottomrule

\end{tabular}
\caption{Test set results for WMH volume classification with varying levels of imbalance for `younger' participants. The number following the method is the percentage of ‘younger’ participants retained in the training and validation sets (1, 10 or 25\%). The number of `older' patients in the training and validation sets is 6305, 566 respectively. Of the `younger' participants in the training and validation sets respectively, 1\% amounts to 64, 6 participants, 10\% amounts to 700, 63 participants, and 25\% amounts to 2101, 188 participants. Here `N/A' indicates that $\Delta L\leq0$ (see \S\ref{cost_index}), so the HEI does not apply.
}\label{abnormal_wmh_varying_younger}
\end{table*}

We again see across-the-board performance improvements that are presumably driven by the predictive models learning to be invariant to confounding features which are irrelevant to the prediction of the target (see the end of \S\ref{sex_prediction}). Here the counterfactuals encourage invariance with respect to the size of the lateral ventricles and sulcii --- ageing features spuriously correlated with the target --- while encouraging dependence on abnormally high signal intensities, the causal explanation for high WMH volume.

\subsubsection{Experiment: collider bias}\label{collider_bias}

Here we evaluate how our six approaches remedy performance and equity impaired by a collider bias \citep{griffith2020collider, watson2019collider, collider_obesity}, where two demographic attributes are correlated with the target variable. Collider bias occurs when data collection is incidentally conditioned on a particular attribute, resulting in a distorted/false correlation between that attribute and a target variable.

We use the experimental setup from the previous section, where the target is WMH volume and one of the confounding demographic attributes is age. We do not evaluate CF-Net because the official implementation \citep{confounder_free} does not support multiple confounders. 

Here we simulate a collider bias by conditioning the generation of two training and validation sets on sex. We begin with all of the `older' males with top quartile WMH volume (see \S\ref{the_data}), then add participants, in equal proportions male, female, `younger', `middle-aged' and `older', until they make up 10\% of the total: this constitutes the first set. We then repeat the process, but now continuing to add participants until 25\% of the total is reached. The test sets are, as always, sampled from the natural distribution.

In the 10\% set the sample Pearson correlation coefficient between sex and the target is $0.84$ with $p < 0.0005$ and between age and the target it is $0.42$ with $p < 0.0005$.
In the 25\% set the sample Pearson correlation coefficient between sex and the target is $0.81$ with $p < 0.0005$ and between age and the target it is $0.39$ with $p < 0.0005$. For reference, the Pearson correlation on the natural UK Biobank distribution between sex and abnormally high WMH volumes is only $0.04$ with $p < 0.0005$.

For each combination of model and collider bias we present (with standard deviation): (1) balanced accuracy; (2) precision and recall for the best- and worst-preforming subpopulation; (3) average balanced accuracy; and (4) our index, HEI. The results are summarised in Table \ref{abnormal_wmh_varying_sex_and_age}. 

\begin{table*}[!t]
\centering
\begin{tabular}{lllllll} \toprule
 Method & Best B-Acc & Best Prec/Recall & Worst B-Acc & Worst Prec/Recall & Avg. B-Acc & HEI  \\ \midrule
 
 ERM, 10\% & 61.6 $\pm$ 2.9 & 56.9/61.6 $\pm$ 2.0/2.9 & 48.6 $\pm$ 0.8 & 48.1/48.6 $\pm$ 0.0/0.8 & 54.9 $\pm$ 1.1 & N/A \\
 
 DRO, 10\% & 60.1 $\pm$ 9.7 & 54.6/60.1 $\pm$ 2.5/3.7 & 52.1 $\pm$ 7.6 & 50.3/52.1 $\pm$ 4.5/3.6 & 54.2 $\pm$ 1.2 & 3.0 \\
 
 ERM-ACF, 10\% & \textbf{70.0} $\pm$ 4.1 & \textbf{69.6}/\textbf{70.0} $\pm$ 3.6/4.1 & 53.1 $\pm$ 3.3 & 52.4/53.1 $\pm$ 4.6/3.3 & \textbf{61.3} $\pm$ 1.4 & 10.5 \\
 
 DRO-ACF, 10\% & 66.6 $\pm$ 3.5 & 59.8/66.6 $\pm$ 2.7/4.5 & 56.2 $\pm$ 2.2 & 53.9/\textbf{56.2} $\pm$ 4.2/2.2 & 59.8 $\pm$ 2.1 & 12.3 \\
 
 ERM-BCF, 10\% & 67.9 $\pm$ 4.0 & 63.4/67.9 $\pm$ 3.8/4.0 & 55.6 $\pm$ 3.6 & \textbf{56.1}/55.6 $\pm$ 0.9/3.6 & 60.7 $\pm$ 2.0 & 12.5 \\
 
 DRO-BCF, 10\% & 66.1 $\pm$ 4.5 & 61.3/66.1 $\pm$ 4.3/4.5 & \textbf{56.3} $\pm$ 3.7 & 55.4/56.3 $\pm$ 0.9/3.7 & 60.3 $\pm$ 1.7 & \textbf{12.8} \\ \midrule

 ERM, 25\% & 63.2 $\pm$ 5.9 & 57.6/63.2 $\pm$ 2.0/4.9 & 48.9 $\pm$ 0.4 & 48.1/48.9 $\pm$ 0.0/0.4 & 55.7 $\pm$ 0.5 & N/A \\
 
 DRO, 25\% & 60.6 $\pm$ 3.6 & 56.0/60.6 $\pm$ 3.0/3.6 & 53.0 $\pm$ 0.2 & 52.2/53.0 $\pm$ 4.2/0.2 & 56.2 $\pm$ 4.1 & 4.6 \\
 
 ERM-ACF, 25\% & 67.9 $\pm$ 1.4 & 70.3/67.9 $\pm$ 0.9/1.4 & 59.6 $\pm$ 3.5 & 55.6/59.6 $\pm$ 1.8/3.5 & 65.0 $\pm$ 0.4 & 19.3 \\
 
 DRO-ACF, 25\% & 64.9 $\pm$ 0.8 & 64.8/64.9 $\pm$ 3.9/0.8 & 49.8 $\pm$ 4.6 & 49.9/49.8 $\pm$ 1.8/4.6 & 58.5 $\pm$ 1.8 & 3.4 \\
 
 ERM-BCF, 25\% & 72.3 $\pm$ 3.7 & 67.6/72.3 $\pm$ 2.4/3.3 & \textbf{60.2} $\pm$ 2.0 & \textbf{56.0}/\textbf{60.2} $\pm$ 4.5/2.0 & 65.5 $\pm$ 1.6 & \textbf{20.4} \\
 
 DRO-BCF, 25\% & \textbf{76.7} $\pm$ 4.3 & \textbf{75.3}/\textbf{76.7} $\pm$ 6.8/3.3 & 59.6 $\pm$ 2.2 & 55.9/59.6 $\pm$ 3.8/2.2 & \textbf{65.7} $\pm$ 1.4 & 19.9 \\ \bottomrule
 
\end{tabular}
\caption{Test set results for WMH volume classification with sex and age as collider variables. Here `N/A' indicates that $\Delta L\leq0$ (see \S\ref{cost_index}), so the HEI does not apply.}\label{abnormal_wmh_varying_sex_and_age}
\end{table*}

Table \ref{abnormal_wmh_varying_sex_and_age} shows an even wider performance gap between best-demographic and worst-demographic performance for both ERM and DRO.

\section{Discussion}\label{discussion}

With arguably the greatest strength of complex modelling --- its individuating power --- comes a critical vulnerability: differential performance across diverse subpopulations in proportion to their representation in the training data \citep{gebru_gender, gender_imbalanced_xray, fairness, fairness_ml}. Where the sampling of the foreground signal of interest is insufficient to reveal its structure, to any conceivable model, adding more data is the only viable solution. But where differential performance arises from the conflation of foreground and incidentally correlated, irrelevant background features, systematic manipulation of the background alone may provide an adequate remedy. Crucially, such manipulation may be informed by data from another domain, executed by models trained under large-scale data regimes infeasible in the target domain. 

Here we demonstrate in the realm of brain imaging a robust method for achieving this by counterfactual synthetic data augmentation constrained to morphological features of the background. We show that this approach can enhance performance on minority subpopulations defined by multiple interacting factors, promoting equity without a cost to the rest of the population, indeed with added benefit (\S \ref{demographically_fair_predictions}). Whereas a closed framework, reliant on redistributing model attention within the domain, such as group distributionally robust optimization \citep{dro_neural_nets}, will generally improve performance in one subpopulation at the cost of degrading it in another, an open framework that transfers knowledge from another domain has the potential to improve equity at no overall cost. 

Seven points of necessity, optimality, generalisability, and scope arise.

First, it should be recognized that in medicine the acquisition of large scale data is often limited by constitution rather than practicality. Neurology in particular is replete with pathological conditions too rare to allow the data scales to which contemporary machine vision architectures are accustomed. Amyotrophic Lateral Sclerosis, for example, is diagnosed in only 670 new patients across the UK annually. Operating with comparatively small scale data is, and always will be, the norm here, making data efficiency an essential aspect of complex analytic methods with real-world ambitions. 

Second, if the necessity for conventional data augmentation, such as geometric transformations, is conceded by its widespread use in contemporary medical imaging models, then its extension to other features to which invariance should be promoted is entirely natural. Note that the biologically-informed augmentation introduced here cannot plausibly be replaced by random non-linear transformations that could superficially mimic it because to achieve adequate disentanglement from correlated factors we need to replicate biologically structured patterns of background variation. 

Third, though non-linear image registration can be used to homogenize images morphologically \citep{evaluation_registration}, it does not provide a practicable means of reducing background contextual entanglement. The regularisation on which robust non-linear registration depends inevitably retains substantial morphological signals as demonstrated by the excellent performance of age regression and sex-classification models on registered data (\S \ref{gans_experiments}). Moreover, whereas augmentation need only be confined to training, a registration-dependent analytic framework requires test data to be transformed into the same space: a task not easily accomplished without interference from foreground pathology.

Fourth, the proposed augmentation strategy does not assume, but is inevitably sensitive to, the preservation of the foreground signal in the act of translation. This is the core rationale for restricting the generator to diffeomorphic morphological deformations that leave tissue intensity signals broadly intact (\S \ref{vbm_experiment}). Where the foreground signal is itself morphologically conveyed, the synthetic mechanism may conceivably distort it. But whether or not such distortion offsets the benefits of augmentation is quantifiable at test time, and will depend on the task and the nature of the pathology. Crucially, the use of a more expressive synthetic model is not necessarily desirable, for the risk of distortion or even erasure of the pathological signal is thereby increased. In situations where the background requires an intensity-based manipulation, an analogous non-morphological generative architecture would be appropriate.

Fifth, though our method is here applied to the promotion of equitable model performance, its use has the potential to harden a model to distributional shift \citet{counterfactual_efficacy} and reduce the risk of underspecification \citep{google_under, gender_imbalanced_xray} by counterfactually exposing it to a wider diversity of plausible foreground-background combinations than the training data alone contains. This should not only lessen model dependence on domain-specific features with poor generalisability, but enable training a model to become cognisant of specific, directed, counterfactually-defined contextualising backgrounds, before they are even encountered in the wild.

Sixth, the ability to learn, transferrably, a characteristic such as age or sex from a set of data will be sensitive to other characteristics, such as the presence of incidental pathology, to the extent to which they interact with it. While we minimize this sensitivity by constraining the expressivity of our synthetic mechanism to modulations of morphology, its magnitude is an empirical question to be answered in any specific modelling scenario by quantifying the fidelity of retrieval of the conditioning characteristic from a separate set of synthesized data. Performing such quantification on the target set of interest may be complicated by the presence of pathology, but the value of the overall augmentation process is in any event ultimately determined by the fidelity of the downstream task, evaluated on held out data.

Finally, casting light on the equity of model performance across subpopulations reveals a pressing need for a quantitative ethical framework that allows formal comparison across both architectures and trained models. Here we build on concepts derived from econometrics to suggest a novel index (\S \ref{cost_index}), the  Holistic Equity Index, that addresses the specific needs of the task, with potential utility in other areas.

\section{Conclusion}

CounterSynth is a novel conditional generative model of diffeomorphic deformations that induces label-driven, biologically plausible changes in volumetric brain images with potential utility in enabling biologically structured counterfactual augmentation. 

We demonstrate by voxel-based morphometry (\S \ref{vbm_experiment}), demographic classification (\S\ref{classification_validation}), and Fr\'{e}chet inception distance (\S\ref{FIDS}) that this model produces anatomical deformations closely replicating the actual demographic morphological  differences observed in UK Biobank data.

Extensive comparative evaluation (\S\ref{demographically_fair_predictions}) on demographically imbalanced tasks with and without confounders further demonstrates that the use of counterfactual augmentation results in state-of-the-art improvements to both overall fidelity and equity of discriminative models, optionally operating in synergy with other fairness methods such as DRO. 

The enviable power of complex modelling in the realm of medical imaging has brought increased focus on the necessity to match performance with equity across heterogeneous populations. Our model and associated analyses cast light on the the problem of equity in modelling brain images, and provide theoretical and practical elements of a framework that will enable researchers and clinicians to tackle it head on.

\section*{Acknowledgments and Funding}

This research has been conducted using the UK Biobank Resource under Application Number 16273. This work is supported by the EPSRC-funded UCL CDT in Medical Imaging (EP/L016478/1), the Wellcome Trust (213038) and NIHR UCLH Biomedical Research Centre.

\bibliographystyle{model2-names} 
\bibliography{refs}

\end{document}